# Protective Self-Adaptive Pruning to Better Compress DNNs

Liang Li, Pengfei Zhao
SRCX, Samsung Electronics
liangli@buaa.edu.cn

## Abstract

Adaptive network pruning approach has recently drawn significant attention due to its excellent capability to identify the importance and redundancy of layers and filters and customize a suitable pruning solution. However, it remains unsatisfactory since current adaptive pruning methods rely mostly on an additional monitor to score layer and filter importance, and thus faces high complexity and weak interpretability. To tackle these issues, we have deeply researched the weight reconstruction process in iterative prune-train process and propose a Protective Self-Adaptive Pruning (PSAP) method. First of all, PSAP can utilize its own information – weight sparsity ratio – to adaptively adjust pruning ratio of layers before each pruning step. Moreover, we propose a protective reconstruction mechanism to prevent important filters from being pruned through supervising gradients and to avoid unrecoverable information loss as well. Our PSAP is handy and explicit because it merely depends on weights and gradients of model itself, instead of requiring an additional monitor as in early works. Experiments on ImageNet and CIFAR-10 also demonstrate its superiority to current works in both accuracy and compression ratio, especially for compressing with a high ratio or pruning from scratch.

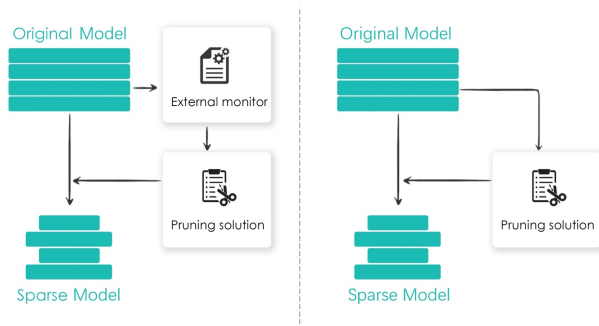

Figure 1: A brief illustration of current adaptive pruning solution (left) and our desired solution (right). Current search-based adaptive pruning methods usually train an external monitor to help make pruning decisions and sample subNets while our design merely relies on model itself.

## Introduction

Network pruning is an effective way to deploy over-parameterized models on mobile devices. Early traditional works (He et al. 2018; He et al. 2019; Li et al. 2016) simply cut off less important filters of each layer with a fixed pruning ratio, and they are likely to cause much accuracy drop and limit model compression ratio since these methods, such as L-SEP (Ding et al. 2021; Han et al. 2015), fail to take importance and redundancy discrepancy among layers into consideration.

More recently, some adaptive solutions emerge to ameliorate the two aforementioned issues and they are capable of allocating an appropriate pruning ratio for each layer. Despite great success in terms of metric evaluations, these solutions remain unsatisfactory due to high complexity and prohibitive computation cost introduced by an external monitor that these solutions rely on to score the importance/redundancy of different layers. For example, Automatic Model Compression (AMC) (He et al. 2018) and NetAdapt (Yang et al. 2018) resort to a reinforcement learning (RL) agent sample suitable pruning ratios from predefined search space, while Soft Channel Pruning (SCP) (Kang and Han 2020) and Dynamic Sparse Training (DST) (Liu et al. 2020) introduce a differentiable mask as a monitor in the training process to determine the pruning ratio of each layer. However, additionally training an excellent RL agent or a differentiable mask is somewhat complex and time-consuming, dissuading individual researchers who do not have sufficient GPU resources. Therefore, a simple and handy adaptive pruning method without external monitors is in great need.

To this end, we focus on a general progressive pruning paradigm called iterative prune-train (IPT) (see Section 2.3 for more details). The attractive point of IPT is that model parameters become sparser after executing pruning and training steps many times. That means some weights finally turn into zeros in this way, which coincides with our compression concept in Fig. 1 - compression along with training process without external monitors. Further research on IPT inspires us to design a novel pruning method. First of all, we notice that weight sparsity ratio (WSR) among layers varies

---

a lot after an IPT process, and weights of unimportant layers always tend to be sparser, vice versa. Therefore, WSR itself is actually a natural reflection of importance and redundancy of layers, and it can be used as an indicator to design our adaptive pruning solution. Secondly, we observe that if some important filters are pruned in the pruning step of IPT, these filters will be correspondingly reconstructed to abnormal maximum values due to some huge gradients (so-called "pulse gradients" here) during the back-propagation of training step. Further research reveals that these pulse gradients are caused by Batch Normalization (BN) (Ioffe and Szegedy 2015) operations, and they probably lead to instability and accuracy deterioration. Thus, there is a demand to resolve the pulse gradient issue and protect these important filters from being pruned.

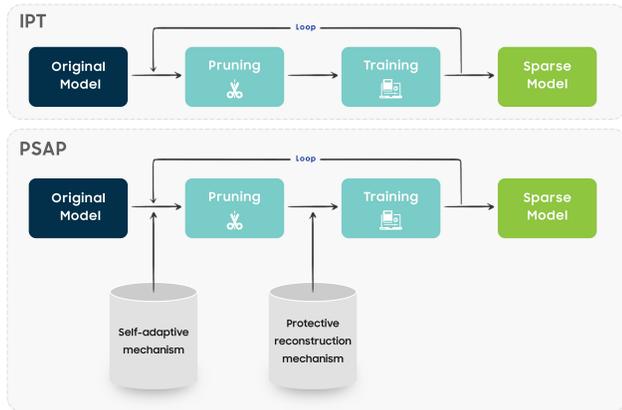

Figure 2: An overview of our proposed PSAP. We introduce two effective mechanisms for better performance based on IPT which iteratively performs pruning and training steps. Specifically, we update the pruning ratios of each layer according to WSRs before the pruning step and repair the pulse gradient issue by using our protective reconstruction mechanism before normal training step.

In this paper, we propose Protective Self-Adaptive Pruning (PSAP), a handy and efficient approach to discover the optimal pruning strategy without external monitors for deep neural networks (DNNs). Specifically, the self-adaptive mechanism features in PSAP, and it has the ability to automatically adjust and update layer pruning ratios before every pruning step, merely based on current WSRs and previous pruning ratios. Moreover, PSAP is also armed with the protective reconstruction mechanism to tackle the pulse gradient issue by pinpointing those pruned filters with pulse gradients and then reloading with previous weights, securing those important filters from being pruned. Fig. 2 is a brief illustration of PSAP.

In summary, the main contributions of this work are summarized below:

- This is the first attempt to notice the value of weight sparsity ratio and take advantage of it to design our self-adaptive pruning mechanism.

- We propose a protective reconstruction mechanism to effectively secure important filters and tackle the pulse gradient issue for training stability and accuracy maintenance.

- Extensive experiments on CIFAR-10 and ImageNet datasets demonstrate that our PSAP achieves great results through pruning from either pre-trained models or scratch and surpasses current state-of-the-art methods in both accuracy and compression ratio.

## Related work

Research on network pruning can be traced back to 1989, Yann (Le Cun, Denker, and Solla 1989) achieves network slimming by removing unimportant weights from the network. In recent years, there has been a lot of literatures about network pruning, and these works can be roughly grouped into three categories by the order of pruning and training step.

**Prune after pre-training.** This type of classic network pruning method generally trains an over-parameterized model first and then cuts off part of less important pre-trained weights based on metrics regarding weight importance, like weight magnitude (Han et al. 2015), restructure error (He, Zhang, and Sun 2017), geometric median in FPGM (He et al. 2019) and correlation (Wang et al. 2019a).

**Prune before training.** Some recent works claim that structure itself matters a lot for model pruning. For example, Lottery Ticket Hypothesis (LTH) (Frankle and Carbin 2019) and Rethinking the value of network pruning (Rethink) (Liu et al. 2019b) prove that comparable or even better performance can also be acquired by directly training a compact model derived from the original model. Instead, Pruning from Scratch (PFS) (Wang et al. 2019b) firstly prunes the network model with randomly initialized weights, and then trains the pruned model from scratch and attains an advanced accuracy.

**Prune during training.** This sort of work hopes to combine the pruning operation with network training to obtain higher accuracy. They generally add sparse constraints for optimization or embed iterative pruning operations into training process to generate a sparse model at last. In general, the desirable sparse structure can be acquired either through sparse regularization learning (Liu et al. 2017; Wen et al. 2016) or by iterative prune-train approaches (He et al. 2018; Mocanu et al. 2018).

### Adaptive pruning

As the neural networks get deeper and the network structure designs become more complex, it is infeasible to determine pruning ratios among layers based on experiences. That is where adaptive pruning approaches emerge and boom recently. For instance, AMC (He et al. 2018) and NetAdapt (Yang et al. 2018) leverage the strength of reinforcement learning to automatically search the design space to make tradeoffs among model size, speed, and accuracy and adaptively explore model structure. G. Ding (Ding et al. 2021) instead employs LSTM to determine pruning policies and combines an SEP attention mechanism to further assist to tailor the model properly. MetaPruning (Liu et al. 2019a) incorporates meta-learning as a controller for pruning. AutoPruner (Luo and Wu 2020) proposes an efficient channel selection layer, to find less important filters automatically in a joint training manner. Additionally, some regularization-based methods like Slimming (Liu et al. 2017) and Sparse Structure Learning (SSL) (Wen et al. 2016) compress the network through sorting the whole weight importance globally according to scale values in Batch Normalization. This regularization-based approach is similar to ours, which also aims to generate a sparse model during training. However, regularization-based methods usually fail to give exact sparse solutions, so an additional fine-tuning is required to enhance the accuracy of pruned model.

### Iterative prune-train IPT

The IPT approach performs pruning and training step iteratively, and it is an iterative and greedy selection procedure to approximately optimize the non-convex problem for finding sparse structures in neural networks. Different from traditional greedy methods, like ThiNet (Luo, Wu, and Lin 2017) and PruneTrain (Lym et al. 2019), that permanently cuts off the weights and will never be restored, IPT can reconstruct part of pruned weights to alleviate accuracy degradation. Sparse Evolutionary Training (SET) (Mocanu et al. 2018) proposes a prune-regrowth procedure that allows the pruned neurons and connections to recover randomly. Dynamic Sparse Reparameterization (DSR) (Mostafa and Wang 2019) uses dynamic parameter reallocation to find the sparse structure. Gradient Hard Thresholding Pursuit (GraHTP) (Yuan, Li, and Zhang 2013) repeatedly performs standard gradient descent followed by a hard truncation operation, instead of a re-initialization operation, and applies it to sparse logistic regression and sparse SVMs learning tasks. Iterative Hard Thresholding (IHT) (Jin et al. 2016) further extends this approach to DNNs and proposes an iterative hard thresholding method that firstly performs hard thresholding to set connections with small magnitudes to zero and then reactivates frozen weights during gradient update. Soft Filter Pruning (SFP) (He et al. 2018) introduces a soft filter pruning approach to softly prune filters in order to maintain the model capacity for better performance. Another work (He et al. 2020) further proposes an asymptotic soft filter pruning (ASFP) method as a refinement of SFP, and it allows pruned filters to be updated and asymptotically adjusts the pruning ratio to approach the predetermined maximum value. However, these works are not satisfactory solutions because they require manually setting pruning hyper-parameters beforehand based on expert experiences.

## Method

This part presents our proposed method in detail, beginning with explicit description of our proposed PSAP, followed by self-adaptive mechanism and protective reconstruction mechanism, and ended with its holistic algorithm procedure.

### Protective self-adaptive pruning (PSAP)

For simplicity, we just discuss the situation with the batch size of 1. We denote the training sample data as $(X, Y^*)$

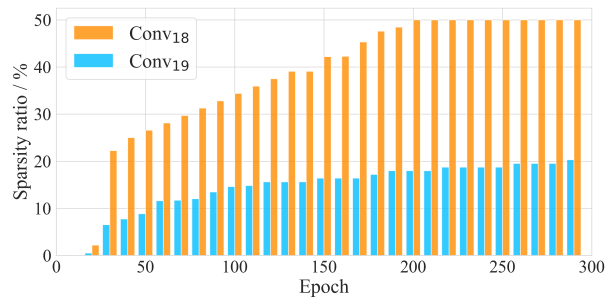

Figure 3: A test for importance discrepancy among layers. For simplification, we select only two convolution layers $Conv_{18}$ and $Conv_{19}$ of ResNet-20 to validate our concept.

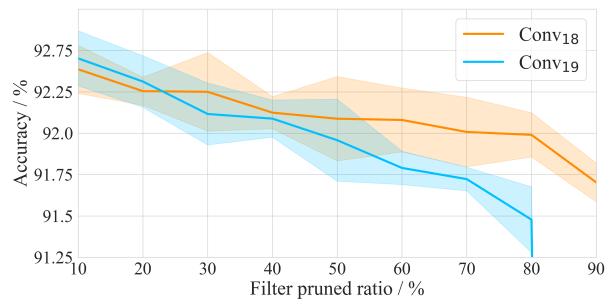

Figure 4: A test to explain the relationship between WSR and layer importance. Note that When the pruning ratio is less than 20%, the result of $Conv_{19}$ is slightly better because of over-fitting.

where $X \in \mathbb{R}^d$ is raw input data and $Y^* \in \{1, 2, \cdots, C\}$ represents the labels with $C$ classes. Now we consider a neural network with totally $L$ layers, denoting the input of $l^{th}$ layer as $X^{(l)}, l \in \{1, 2, \cdots, L\}$. Let $W^{(l)}$ denote the weights of the $l^{th}$ layer and $W = [W^{(1)}, \cdots, W^{(L)}]$. We use $W^*$ for weights after training. Based on the above definitions, a normal optimization through minimizing the loss function without any constraint can be expressed as

$$\min_{W} \left( \mathcal{L}(X, Y^*, W) + \mathcal{R}(W) \right) \quad (1)$$

where $\mathcal{L}(X, Y^*, W)$ is loss function such as cross-entropy loss for image classification task, and $R(W)$ is a non-structured regularization applied on every weight, like $L2$ norm, as weight decay.

In this work, we impose two constraints on this optimization process, and the optimization target is changed to:

$$\min_{W} \left( \mathcal{L}(X, Y^*, W) + \mathcal{R}(W) \right)$$
$$s.t. \ \frac{\|W^*\|_0}{F(W)} \leq \tau \ \& \ \frac{\|W^{(l)}\|_0}{F(W^{(l)})} \geq S_{min} \quad (2)$$

where $A(\cdot)$ is the number of relevant learnable parameters. $\|W^*\|_0/F(W)$ and $\|W^{(l)}\|_0/F(W^{(l)})$ denote the final sparsity ratio and WSR of $l^{th}$ layer, respectively. $\tau$ is the tar-get compression ratio and $S_{min}$ is the minimal compression ratio for all layers.

Algorithm 1: Algorithm procedure of PSAP
**Input**: Data $X$, Label $Y^*$, Model $W$.
**Parameter**: Training and pruning-related parameters.
**Output**: Compressed model.
1: Given hyper-parameters, such as the sparsity increment $\delta$, total and minimal compression ratio $\tau$, $S_{min}$.
2: **Stage 1. Adative solution searching :**
3: **while** Current compression ratio $\leq \tau$ **do**
4:   **if** search_epoch == 1 **then**
5:     Set the initial pruning ratio of each layer to 0.1.
6:   **else**
7:     Update current pruning ratios with the strategy of **self-adaptive mechanism**.
8:   **end if**
9:   Pruning selected filters with current pruning ratios.
10:   Reload important filter weights according to **protective reconstruction mechanism**.
11:   Forward and back-propagation.
12: **end while**
13: **Stage 2. Fine-tuning :**
14: **while** train_epoch $\leq MAX\_EPOCHS$ **do**
15:   Forward and back-propagation based on the searched model.
16: **end while**
17: **return** Compressed model.

Under these two constraints, a sparse model certainly can be achieved by iteratively performing pruning and training steps like in Original IPT. But it cannot be a optimal solution, since it lacks guides on how to adaptively allocate pruning ratio for each layer and on how to deal with the pulse gradient issue. Therefore, PSAP introduces two aforementioned mechanisms into origical IPT to stably and efficiently compress DNNs as shown in Algorithm 1. Self-adaptive mechanism aims to alter the pruning ratio of each layer according to its corresponding WSR value before the pruning step. And protective reconstruction mechanism repairs pulse gradient issue to reconstruct weights by pinpointing the indices and reloading with previous weights before the training step.

**Self-adaptive mechanism**

It is commonsense that importance and redundancy among layers of DNNs differ greatly from each other, and therefore setting an adaptive pruning ratio for each layer is an ideal solution. But how to efficiently identify the maximum value of pruning ratio without much accuracy drop remains unsolved.

We propose a handy, self-adaptive mechanism that regards WSR as an innate indicator to monitor and score layer importance and redundancy along with the model training process. This innate indicator saves a lot of energy and resources while external monitors, like RL agent in AMC, may require extra GPU days to train.

In addition, we conceive some experiments to demonstrate 1) importance discrepancy among layers, 2) the tight relation between WSR and layer importance. Here we refactor layer importance as accuracy for simplification. Fig. 3 shows the WSR changes of layer $Conv_{18}$ and $Conv_{19}$ in ResNet-20 along with training epochs. Note that we set the same pruning ratio 50% for these two layers throughout the IPT training process. It is obvious that $Conv_{18}$ is always sparser than $Conv_{19}$, which means the layer $Conv_{19}$ contains more effective information that cannot be pruned and it is more important than $Conv_{18}$. Fig. 4 depicts the accuracy drop along with the increasing pruning ratio of $Conv_{18}$ and $Conv_{19}$. What we highlight here is that the two curves (shown in blue and orange) represent two control experiments. The orange describes that we only prune the layer $Conv_{18}$ with an increasing pruning ratio while other layers remain unchanged, and we only prune the layer $Conv_{19}$ for the blue one. The accuracy of the blue curve drops a lot, which means the layer $Conv_{19}$ is more important and hard to be compressed. Therefore, it can be concluded that there is a tight relation between WSR and layer importance and we can utilize WSR to design our self-adaptive mechanism.

Let $W^{(l)}$ denote the weights of the $l^{th}$ layer, and self-adaptive mechanism consists of the following three steps:

- Calculate the current WSR of each layer $s^{(l)} = 1 - \frac{\|W^{(l)}\|_0}{F(W^{(l)})}$

- Update current pruning ratio $k^{(l)}$ of each layer with a constant increment $\delta$ according to WSR $s^{(l)}$

$$k^{(l)} = \begin{cases} s^{(l)} + \delta, & if \ s^{(l)} \leq k^{(l)} \\ s^{(l)}, & else \end{cases} \quad (3)$$

- Obtain the filter indices with smaller L2 norm values according to their respective pruning ratios

**Protective reconstruction mechanism**

Since we obtain the filter indices after self-adaptive mechanism, the common approach usually directly cuts off those filters with given indices in the pruning step of PSAP as shown in Fig. 2. However, this smaller-norm-less-informative assumption is not always right and unconditional which has been discussed in (Ye et al. 2018). We notice that it would lead to pulse gradients and then reconstruct weights of corresponding filters to maximum values during the back-propagation of IPT, and finally result in training instability and accuracy degradation. The pulse gradient issue is caused by gradient propagation of the following Batch Normalizaiton(BN) (Ioffe and Szegedy 2015) layer since the midterm $1/\sqrt{\sigma^2 + \varepsilon}$ of BN would produce a singularity when the relevant filters are set to zero (refer to Appendix for more details).

Our protective reconstruction mechanism is capable to handle this issue. It has the ability to pinpoint those important filters relevant to pulse gradients and reload with its previous weights. The specific procedure is illustrated in Listing 1.

Listing 1: Protective reconstruction procedure
```
1  # Back up the previous weights.
2  backup_weights()
3
4  # Obtain pruned model.
5  do_mask()
6
7  # Obtain gradients through one gradient
          descent on a mini-batch data.
8  one_gradient_denscent()
9
10 for index, weight in enum(parameters) :
11   if index in mask_index :
12     norm = L2_norm(weight)
13     norm_mean = norm.mean()
14
15     # pinpoint abnormal filters
16     while i < len(norm) :
17       if norm[i] > norm_mean :
18         abnm_filters.append(i)
19
20     # reload with previous weights
21     while k < len(abnm_filters) :
22       weight[k] = backup_weights[k]
```

## Experiments

### 1 Benchmark

1) Datasets. We use the CIFAR-10 (Krizhevsky 2009) and ImageNet (Krizhevsky, Sutskever, and Hinton 2012) datasets in this work. The CIFAR-10 dataset consists of 50,000 training images and 10,000 test images of ten classes. The ILSVRC-2012 dataset is pretty large, composed of 1.28 million training images and 50,000 validation images of 1,000 categories.

2) Networks. We demonstrate the performance of our method by pruning two challenging and widespread networks, ResNet (He et al. 2016a) and MobileNetV2 (Sandler et al. 2018). ResNet series has been widely used in many deep learning scenarios as the backbone of network, and it is meaningful to make some progresses for a lightweight ResNet. MobileNetV2, as the representatives of small-size networks, inclines more to edge devices. Better pruning performance on these two classic and challenging networks will definitely show the superiority and effectiveness of our method.

### 2 Implemetation Details

The hyper-parameters of our method can be grouped into two parts, basic training parameters and pruning-related hyper-parameters. We execute all the experiments on one Tesla V100 using Pytorch as backend and follow the default training parameter setting in (He et al. 2016b). For the experiments on CIFAR-10 dataset, we use the same setting as SFP (He et al. 2018), the learning rate is initially set to 0.05 and descends with the same decay path as (Zagoruyko and Komodakis 2016), and the batch size is 128. We employ SGD optimizer with a weight decay of 5e-4 and a momentum of 0.9. As for the ImageNet dataset, we compress ResNet-50 with an initial learning rate of 0.2 and MobileNetv2 with learning rate of 0.1, and the learning rate descends in a linear manner. When it comes to pruning-related hyper-parameters, we initially set the pruning ratio $k^{(l)}$ to 0.1 for each layer and then dynamically adjust it according to the self-adaptive mechanism where the constant increment of pruning ratio $\delta$ is 0.2 throughout all the experiments. The minimal sparsity ratio of each layer $S_{min}$ is set to 0, meaning that the entire layer can be removed.

The PSAP training process consists of two phases: searching and fine-tuning phase. In searching phase, we always comply with the procedure illustrated in Fig. 2 until the current total pruning ratio attains the overall predefined compression ratio. In fine-tuning phase, we only perform a normal training with the fixed pruning ratios of $L$ layers obtained in searching phase. For training epoch setting, it depends on the desire compression ratio, target network, and datasets. For instance, if we plan to prune ResNet-56 on CIFAR-10 with a compression ratio of 0.2, 30 epochs are required for searching phase and 100 epochs for fine-tuning phase. But if there is an experiment for ResNet-50 on ImageNet with a compression ratio of 0.75, 250 epochs are required for searching and 200 for fine-tuning.

### 3 Comparison with SotA approaches

Overall, our proposed PSAP surpasses the SotA approaches by a clear margin in both accuracy and compression ratio on two datasets. Note that all the results of SotA methods are the best ones retrieved from their papers. In addition, since the base accuracy of these methods are different, we utilize accuracy drop as the primary metric and accuracy itself for the secondary.

**ResNet-56 on CIFAR-10.** As shown in Table 1, our approach achieves an accuracy of 93.74% with an increment of 0.15% by pruning about 50% FLOPs while others face some accuracy drops. It is true that DCP (Zhuang et al. 2018) is prominent and obtains 93.81% in this task, but this high accuracy attributes more to its higher baseline 93.80%. Furthermore, our PSAP also performs well when pruning with higher ratios. For instance, it still has the ability to retain a high classification accuracy of 90.76% with an pruning ratio of almost 80%, which is pretty hard for other methods.

| Method | Base(%) | Acc.(%) | Acc.↓(%) | FLOPs↓ |
|---|---|---|---|---|
| SFP | 93.59 | 93.35 | 0.24 | 52.6% |
| FPGM | 93.59 | 93.49 | 0.10 | 52.6% |
| AMC | 92.8 | 91.9 | 0.9 | 50.0% |
| PFS | 93.23 | 93.05 | 0.18 | 50.0% |
| SCP | 93.69 | 93.23 | 0.46 | 54.3% |
| DCP | 93.80 | 93.81 | -0.01 | 47.1% |
| Ours | 93.59 | **93.74±0.24** | **-0.15** | 48.6% |
|  |  | 93.33±0.19 | 0.26 | 62.1% |
|  |  | 92.89±0.13 | 0.70 | 69.0% |
|  |  | 90.76±0.24 | 2.83 | 79.5% |

Table 1: ResNet-56 on CIFAR-10 dataset. "Base(%)" stands for the top-1 accuracy of pretrained models. "Acc.(%)" and "Acc.↓" represent the top-1 accuracy of pruned models and its relevant accuracy drop compare with baseline, respectively. "FLOPs↓" denotes the pruned ratio of FLOPs.

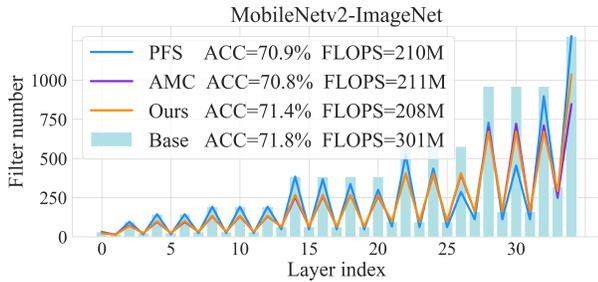

Figure 5: Visualization of three adaptive pruning methods, including PFS (Wang et al. 2019b), AMC (He et al. 2018) and our PSAP.

**ResNet-50 on ImageNet.** Table 2 reveals that our method far outperforms other advanced methods. Our PSAP achieves a salient accuracy of 76.83% with an increment of 0.18% by cutting off 36.6% FLOPs while most of other methods stay 75% or so. For fair comparison in higher pruning ratios, we compare the proposed method with PFS from three FLOPs settings. It is obvious that our PSAP performs better than PFS, especially at higher compression ratios. For instance, we attain a noticeable score of 73.54% on ImageNet at 1.0G FLOPs, better than 72.80% of PFS.

**MobileNetV2 on ImageNet.** As designed for edge devices, MobileNetV2 is so compact that there is a little space for compression, bringing great challenges to this task. Note that we have to ensure the index consistency of pruned filters for point-wise and depth-wise convolutions. Table 3 shows that we obtain the best performance of 71.4% with the smallest accuracy drop of 0.7% under the condition of pruning ratio of about 30%. In addition, we compare our allocation solution with that of PFS and AMC by visualizing their remained filter numbers of each layer in Fig. 5. As you can see, the allocation solution of AMC is very similar to ours but different from PFS. This is because both AMC and PSAP prune from pre-trained model, but PFS prunes form scratch with randomly initialized weights which has more potential paths to explore. Our method outperforms these two advanced methods, benefiting a lot from our protective mechanism for important filters.

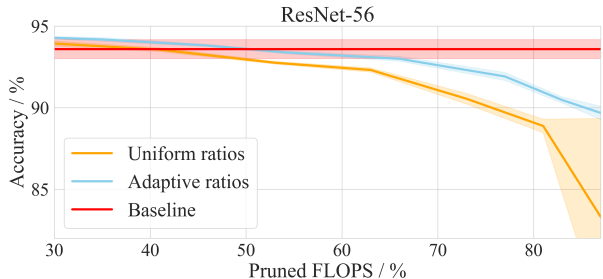

Figure 6: Qualitative analysis of self-adaptive mechanism. The uniform results here are from SFP, and the adaptive is from our PSAP without the protective reconstruction mechanism.

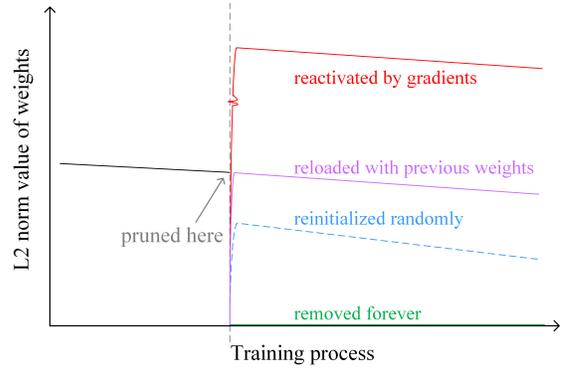

Figure 7: Impacts on weights of a typical filter through four distinct restruction paths. After pruning, the $L2$ norm value of weights changes a lot except through reloading.

| Method | Base(%) | Acc.(%) | Acc.↓(%) | FLOPs | FLOPs↓ |
|---|---|---|---|---|---|
| SFP | 76.15 | 74.61 | 1.54 | 2.4G | 41.8% |
| ASFP | 76.15 | 75.53 | 0.62 | 2.4G | 41.8% |
| L-SEP | 76.12 | 75.22 | 0.90 | 2.3G | 43.0% |
| SCP | 75.89 | 75.27 | 0.62 | 1.9G | 54.3% |
| DCP | 76.01 | 74.95 | 1.06 | 1.8G | 55.0% |
| PFS | 77.20 | 76.70 | 0.50 | 3.0G | 26.3% |
|  |  | 75.60 | 1.60 | 2.0G | 51.2% |
|  |  | 72.80 | 4.40 | 1.0G | 75.6% |
| Ours | 76.65 | **76.83** | **-0.18** | 2.6G | 36.6% |
|  |  | **76.37** | **0.28** | 2.1G | 48.8% |
|  |  | **73.54** | **3.11** | 1.0G | 75.6% |

Table 2: ResNet-50 on ImageNet dataset.

## 4 Ablation study

It is worth mentioning that the accuracy enhancement and compression ratio extension benefit from our self-adaptive mechanism and protective reconstruction mechanism. Quantitative and qualitative analysis about these two mechanisms will be given in this part.

**Quantitative analysis.** To verify the effectiveness of the proposed two mechanisms, we conduct four ResNet-56 control experiments with a compression ratio of 50% on CIFAR-10. As shown in Table 4, compared with pure IPT, either self-adaptive mechanism or protective reconstruction mechanism is effective, and PSAP with two mechanisms achieves the accuracy of 93.74%, higher than IPT by nearly 1%.

**Qualitative analysis of self-adaptive mechanism.** To demonstrate its superiority on adaptivity, we conceive two experiments to prune the ResNet models with a series of different compression targets on CIFAR-10 dataset. We set one experiment (PSAP, w/o PR) with our self-adaptive mechanism to update its pruning ratio during training process while the other (SFP) fixes all layers with a uniform pruning ratio. We show the training accuracy results in Fig. 6 It can be seen that the curve without adaptive mechanism suffers from sudden drops, especially when the pruned FLOPs exceeds 50%. Given a large compression ratio requirement, the PSAP can still remain sound and safe with a pretty satisfactory accura-cy, but SFP fails to do so, which in this case results in the loss of much important information. In addition, we observe that our PSAP can explore and identify the pruning preferences according to model structure. For instance, PSAP prefers to allocate a smaller pruning ratio to the last convolution layer of ResBlock which is connected to other blocks. This preference coincides with early manual solutions based on expert experiences for largely maintaining the network capacity.

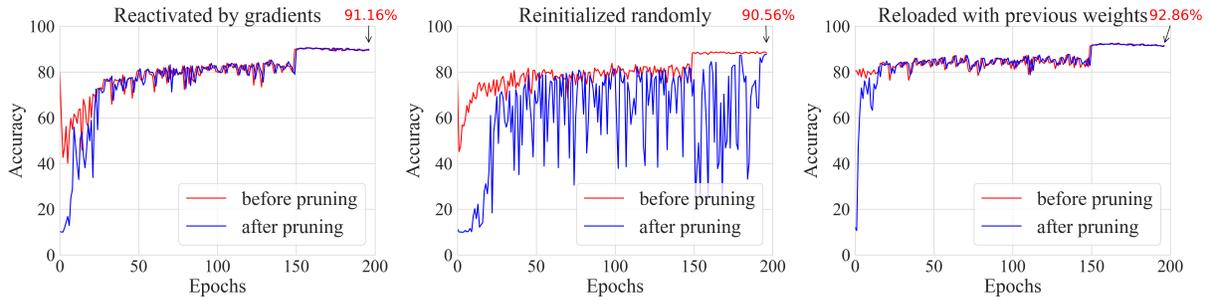

Figure 8: Qualitative analysis of three restruction paths. The red and blue line represent the accuracy of the model before and after pruning, respectively.

| Method | Base(%) | Acc.(%) | Acc.↓(%) | FLOPs | FLOPs↓ |
|---|---|---|---|---|---|
| MetaPruning | 72.1 | 71.2 | 0.9 | 217M | 28.7% |
| AMC | 71.8 | 70.8 | 1.0 | 210M | 30.0% |
| PFS | 72.1 | 70.9 | 1.2 | 210M | 30.0% |
| AutoPruner | 72.2 | 71.2 | 1.0 | 208M | 30.5% |
| Ours | 72.1 | **71.4** | **0.7** | **208M** | **31.0%** |

Table 3: MobileNetV2 on ImageNet dataset.

**Qualitative analysis of protective reconstruction mechanism.** We further study the effectiveness of three weight reconstruction manners on IPT (PSAP, w/o SA) by pruning ResNet-56 on CIFAR-10 with a compression ratio of 50%, involving reactivating via gradients, reinitializing randomly, and reloading with previous weights as shown in Fig. 7. Results in Fig. 8 reveal that the third manner of reloading with previous weights is the most stable one which achieves an excellent accuracy of 92.86%. The manner of reinitializing randomly shows dramatic fluctuations throughout the whole training process and it has a negative impact on accuracy.

**Discussion on pruning from scratch.** It is necessary to discuss the approach of pruning from scratch because it is efficient and productive and does not require setting aside extra time and resources to prepare pre-trained models with high accuracy. Also, we even cannot access the officially available pre-trained models in some circumstances especially when facing prohibitive patent fee or network failures. Therefore, it is promising to explore this path. Our proposed PSAP supports to prune from scratch with randomly initialized weights and we provide the experimental results on CIFAR-10 and ImageNet datasets in Table 5. We notice that pruning from scratch could obtain comparable or even better accuracy than current SotA methods, though slightly lower than the results of our PSAP from pre-trained models. For example, PSAP from scratch can achieve 75.91% and 70.80% on ImageNet task by cutting off 48.8% FLOPs of ResNet-50 and 31.0% of MobileNetV2, respectively.

| Type | Self-adaptive | Protective restruction | Acc.(%) |
|---|---|---|---|
| Pure IPT | × | × | 92.75 |
| PSAP (w/o PR) | √ | × | 93.39 |
| PSAP (w/o SA) | × | √ | 92.86 |
| PSAP (Ours) | √ | √ | **93.74** |

Table 4: Ablation experiments regarding two mechanisms. We use the abbreviations "SA" and "PR" to represent our self-adaptive mechanism and protective reconstruction mechanism.

| | Acc.(%) | FLOPs | FLOPs(%) |
|---|---|---|---|
| ResNet-56 (CIFAR-10) | 93.54±0.10 | — | 45.0% |
| | 93.14±0.24 | — | 56.1% |
| | 92.37±0.15 | — | 69.1% |
| | 91.09±0.28 | — | 81.0% |
| ResNet-50 | 76.54 | 2.5G | 39.1% |
| | 75.91 | 2.1G | 48.8% |
| | 73.42 | 1.0G | 75.6% |
| MobileNetV2 | 70.8 | 208M | 31.0% |

Table 5: Results of pruning from scratch. We prune ResNet-50 and MobileNetV2 on ImageNet dataset.

## Conclusions

In this paper, we abandon search-based adaptive approaches with external monitors and propose a protective self-adaptive pruning solution. Firstly, we verify the value of weight sparsity ratios to dynamically adjust the pruning ratio of each layer beforehand. Secondly, we notice the pulse gradient issue in back-propagation process and then effectively tackle it by locating relevant filter indices and reloading with their previous weights. The experiments on CIFAR-10 and ImageNet datasets show that our method outstrips other current advanced methods in terms of both accuracy and compression ratio. Furthermore, the ability to remain an excellent accuracy under extreme conditions, such as pruning with a large ratios or pruning from random initialized models, demonstrates the superiority of our method.

# Protective Self-Adaptive Pruning to Better Compress DNNs

## Appendix

**Details about protective reconstruction mechanism**

As discussed in the body part, there exists a pulse gradient issue when some important filters are pruned. More generally, the gradients relevant to those pruned filters change dramatically and part of the zero filters can be reconstructed to non-zero values after back-propagation process. Here come explanations.

For simplicity, we just discuss the situation with the batch size of 1. We denote the training sample data as $(X, Y)$ where $X \in \mathbb{R}^d$ is raw input data and $Y$ is output. Now we consider a neural network with totally $L$ layers, denoting the input of $l^{th}$ layer as $X^{(l)}, l \in \{1, 2, \cdots, L\}$, and the output as $Y^{(l)}$. Let $W^{(l)}$ denote the weights of the $l^{th}$ layer and $W = [W^{(1)}, \cdots, W^{(L)}]$. Based on the above definitions, the output $Y^{(l)}$ of the $l^{th}$ convolution layer can be expressed as

$$Y^{(l)} = W^{(l)} * X^{(l)}, \; l = 1, \cdots, L \quad (1)$$

After Batch Normalization (BN) (Ioffe and Szegedy 2015), its output $Z^{(l)}$ is changed to

$$Z^{(l)} = BN^{(l)}_{\gamma, \beta}\left(Y^{(l)}\right) \quad (2)$$

where $\gamma$ and $\beta$ are learnable scale and shift parameters, respectively.

For the $i^{th}$ output of a typical filter $z_i^{(l)}$ in $l^{th}$ layer, it can be written as

$$z_i^{(l)} = \beta_i^{(l)} + \gamma_i^{(l)} \frac{\left(y_i^{(l)} - \mu\left(y_i^{(l)}\right)\right)}{\sqrt{\sigma\left(y_i^{(l)}\right)^2 + \varepsilon}} \quad (3)$$

where $y_i^{(l)}$ is the corresponding $i^{th}$ filter output of $l^{th}$ convolution layer. $\mu\left(y_i^{(l)}\right)$ and $\sigma\left(y_i^{(l)}\right)^2$ are the mean and variance of $y_i^{(l)}$. For conciseness, we abbreviate them to $\mu, \sigma^2$.

Now, we formulate the gradient of the $i^{th}$ filter $W_i^{(l)}$ according to the rule of chain derivation.

$$\frac{\partial L}{\partial W_i^{(l)}} = \frac{\partial L}{\partial z_i^{(l)}} \frac{\partial z_i^{(l)}}{\partial y_i^{(l)}} \frac{\partial y_i^{(l)}}{\partial W_i^{(l)}} \quad (4)$$

where $\frac{\partial z_i^{(l)}}{\partial y_i^{(l)}}$ is the gradient item of BN, it can be factorized as

$$\begin{aligned}\frac{\partial z_i^{(l)}}{\partial y_i^{(l)}} &= \gamma_i^{(l)} \left( \frac{\partial z_i^{(l)}}{\partial y_i^{(l)}} \frac{\partial y_i^{(l)}}{\partial y_i^{(l)}} + \frac{\partial z_i^{(l)}}{\partial \mu} \frac{\partial \mu}{\partial y_i^{(l)}} + \frac{\partial z_i^{(l)}}{\partial \sigma^2} \frac{\partial \sigma^2}{\partial y_i^{(l)}} \right) \\ &= \gamma_i^{(l)} \left( \frac{1}{\sqrt{\sigma^2+\varepsilon}} - \frac{1}{\sqrt{\sigma^2+\varepsilon}} \frac{\partial \mu}{\partial y_i^{(l)}} - \frac{y_i^{(l)}-\mu}{2(\sigma^2+\varepsilon)^{\frac{3}{2}}} \frac{\partial \sigma^2}{\partial y_i^{(l)}} \right) \end{aligned} \quad (5)$$

where $\varepsilon$ denotes a minimum.

If we set weights of the j$^{th}$ filter $W_j^{(l)}$ to zero, the importance score $\left\|W_j^{(l)}\right\|_2$ naturally is also zero. And now we get $y_j^{(l)} = \mathrm{O}$ and $\mu = \sigma^2 = 0$. Thus, the gradient update for normal filters and zeros filters differs a lot, and the output of $\frac{\partial z_i^{(l)}}{\partial y_i^{(l)}}$ could be a maximum $INF$. But the network is still able to converge eventually because the gradient value is actually limited due to the gradient clip operation in real training process. Then the gradient of $W_i^{(l)}$ can be described as

$$\frac{\partial L}{\partial \left(W_i^{(l)}\right)^*} = \begin{cases} \frac{\partial L}{\partial W_i^{(l)}}, & i \neq j \\ \frac{\partial L}{\partial z^{(l)}} \cdot \frac{\partial y^{(l)}}{\partial W_i^{(l)}} \cdot INF, & i = j \end{cases} \quad (6)$$

We point here that it is possible to generate a relatively large pulse gradient corresponding to a specific zero filter due to joint effects of $\frac{\partial L}{\partial z^{(l)}} \cdot \frac{\partial y^{(l)}}{\partial W_i^{(l)}} \cdot INF$. Specifically, when those important filters are pruned in pruning steps, the impact on loss function is pretty significant, and then correspondingly the value of $\frac{\partial L}{\partial z^{(l)}}$ is also great, leading to training instability and accuracy drop. Our protective reconstruction mechanism is capable of tackle this problem by locating the indices of those important filters which have been pruned and then reloading with their previous weights before training steps.

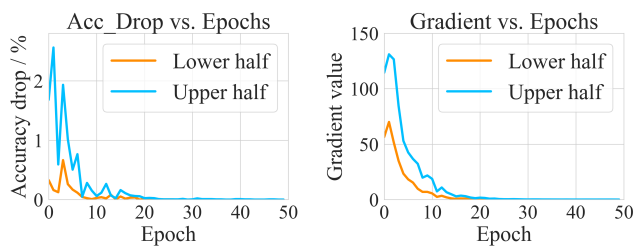

Figure 1: Relation between gradient and accuracy.

**Relation between gradient and accuracy**

As illustrated in Fig. 1, we conceive an experiment by cutting off half weights either with lower importance or with upper one in the training process. Results reveal that pulse gradients are tightly correlated with those more important filters that can cause huge damage to final accuracy once they are set to zero. It is clear that a large gradient always corresponds to a big accuracy drop.